\documentclass[letterpaper, 10 pt, conference]{ieeeconf}  % Comment this line out if you need a4paper

\IEEEoverridecommandlockouts                              % This command is only needed if 
\usepackage[table]{xcolor}
\usepackage{amsmath,amssymb,amsfonts}
\usepackage{stfloats}
\usepackage{multicol}
\usepackage{gensymb}
\usepackage{siunitx}
\usepackage{url}
\usepackage{graphicx}
\usepackage{stackengine}
\usepackage{subcaption}
\usepackage{mwe}
\usepackage{multirow}
\usepackage{hyperref}
\usepackage{lastpage}
\usepackage{fancyhdr}

\hypersetup{
    colorlinks=true,
    linkcolor=blue,
    filecolor=magenta,      
    urlcolor=blue,
    pdftitle={Overleaf Example},
    pdfpagemode=FullScreen,
    }
\usepackage[font={small}]{caption}
\usepackage[font=bf]{caption}
\usepackage{mathtools}
\usepackage{subcaption}
\usepackage[justification=centering]{caption}
\usepackage{soul}

\newcommand{\classname}[1]{\texttt{#1}}
\newcommand{\RN}[1]{%
  \textup{\uppercase\expandafter{\romannumeral#1}}%
}
\title{\LARGE \bf
OpenCDA: An Open Cooperative Driving Automation Framework Integrated with Co-Simulation
}

\author{Runsheng Xu$^{1}$, Yi Guo$^{2}$, Xu Han$^{1}$, Xin Xia$^{1}$, Hao Xiang$^{1}$, Jiaqi Ma$^{1*}$ % <-this % stops a space
\thanks{$^{1}$Runsheng Xu, Xu Han, Xin Xia,  Hao Xiang, Jiaqi Ma are with UCLA Mobility Lab, University of California, Los Angeles (UCLA)}%
\thanks{$^{2}$Yi Guo is with University of Cincinnati}%
\thanks{$^{*}$Corresponding: jiaqima@ucla.edu}%
}

\begin{document}

 \maketitle
\thispagestyle{plain}
\pagestyle{plain}

%%%%%%%%%%%%%%%%%%%%%%%%%%%%%%%%%%%%%%%%%%%%%%%%%%%%%%%%%%%%%%%%%%%%%%%%%%%%%%%%
\begin{abstract}
Although Cooperative Driving Automation (CDA) has attracted considerable attention in recent years,  there remain numerous open challenges in this field. The gap between existing simulation platforms that mainly concentrate on single-vehicle intelligence and CDA development is one of the critical barriers, as it inhibits researchers from validating and comparing different CDA algorithms conveniently. To this end, we propose OpenCDA, a generalized framework and tool for developing and testing CDA systems in simulation. Specifically, OpenCDA is composed of three major components: a co-simulation platform with simulators of different purposes and resolutions, a full-stack prototype cooperative driving system, and a scenario manager. Through the interactions of these three components, our framework offers a straightforward way for researchers to test different CDA algorithms at both levels of traffic and individual autonomy. More importantly, OpenCDA is highly modularized and installed with benchmark algorithms and test cases. Users can conveniently replace any default module with customized algorithms and use other default modules of the CDA platform to perform evaluations of the effectiveness of new functionalities in enhancing the overall CDA performance. An example of platooning implementation is used to illustrate the framework's capability for CDA research. The codes of OpenCDA are available at the \href{https://github.com/ucla-mobility/OpenCDA }{UCLA Mobility Lab GitHub page}. 

\end{abstract}

%%%%%%%%%%%%%%%%%%%%%%%%%%%%%%%%%%%%%%%%%%%%%%%%%%%%%%%%%%%%%%%%%%%%%%%%%%%%%%%%
\section{INTRODUCTION}
By leveraging cutting-edge technologies to circumvent traditional infrastructure enhancement constraints, Intelligent Transportation Systems (ITS) are reshaping transportation and have demonstrated a tremendous potential to boost the transportation system management, operations, safety, and efficiency. One of the essential sub-fields in ITS is Cooperative Driving Automation (CDA), which is defined in SAE J3216~\cite{SAE} and refers to vehicle-highway automation that uses Machine-to-Machine communication to enable cooperation among two or more entities (e.g., vehicles, pedestrians, infrastructure components) with capable communication technologies. By enabling the status-sharing, intent-sharing, and maneuver cooperation between entities, the traffic efficiency, energy consumption, and safety of the driving can be significantly improved~\cite{6489852}. Developed by the Federal Highway Administration (FHWA), the CARMA Program \,\cite{lochrane2020carma} is a leading research program on CDA, leveraging emerging capabilities in automation and cooperation to advance transportation systems management and operations (TSMO) strategies.

Although CDA  has been an active field in recent years, it is still in its infancy. One of the major barriers to the development of CDA is the high cost and potential safety issues to conduct field experiments as they usually require multiple expensive connected automated vehicles (CAVs) and extra-large testing space~\cite{hyldmar2019}. One approach to facilitating experimental research with minimum cost is to prototype and validate the CDA algorithms in a simulated environment. However, existing simulation platforms featured with full-stack software development of autonomous driving provided limited supports for CDA capabilities. As far as we know, there is no existing open-source (or commercial) tool dedicated to CDA by featuring both traffic and vehicles with full CDA vehicle software pipeline. As a result, it becomes very challenging to find an easy and flexible way for researchers to deploy, validate and compare the impact of different CDA algorithms on the dynamic driving tasks of CAVs in simulation. Most of the research uses ad-hoc simulation capabilities with different qualities, making the algorithmic and functional performance not comparable between studies.

To overcome such challenges, we introduce OpenCDA, a generalized open-source framework integrated with co-simulation for CDA research. OpenCDA provides a full-stack CDA software in simulation that contains the common self-driving modules composed of sensing, computation, and actuation capabilities, and cooperative features as defined in SAE J3216 (e.g., vehicular communication, information sharing, agreements seeking). OpenCDA is developed purely in Python~\cite{vanRossum1991InteractivelyTR} for fast prototyping. Built upon these basic modules, OpenCDA supports a range of common cooperative driving applications, such as platooning, cooperative perception, cooperative merge, and speed harmonization. More importantly, OpenCDA offers a scenario database that includes various standard scenarios for testing of different cooperative driving applications as benchmarks. Users can easily replace any default modules in OpenCDA with their designs and test them in the supplied scenarios. If users desire to produce their scenarios, our framework also provides simple APIs to support such customization. We select CARLA~\cite{Dosovitskiy17} and SUMO~\cite{SUMO2018} to render the environment, simulate the vehicle dynamics, and generate the traffic flow. Since our framework is designed with high flexibility, it can also be extended to integrate additional simulators, such as communication simulators (ns-3~\cite{ns32017}) and vehicle dynamics simulators (e.g., CARSim~\cite{carsim}).

The key features of OpenCDA can be summarized as \textbf{IFMBC}:
\begin{itemize}
    \item \textbf{I}ntegration: OpenCDA integrates CARLA and SUMO together for realistic scene rendering, vehicle modeling and traffic simulation.
    \item \textbf{F}ull-stack prototype CDA Platform in Simulation: OpenCDA provides a simple prototype automated driving and cooperative driving platform, all in Python, that contains perception, localization, planning, control, and V2X communication modules.
    \item \textbf{M}odularity: OpenCDA is highly modularized, enabling users to conveniently replace any default algorithms or protocols with their own customzied design. 
    \item \textbf{B}enchmark: OpenCDA offers benchmark testing scenarios, state-of-the-art benchmark algorithms for all modules, benchmark testing road maps, and benchmark evaluation metrics.
    \item \textbf{C}onnectivity and \textbf{C}ooperation: OpenCDA supports various levels and categories of cooperation between CAVs in simulation. This differentiates OpenCDA from other single vehicle simulation tools.

\end{itemize}

The paper is organized as follows. Section \RN{2} will review existing frameworks aiming to support cooperative driving automation and different cooperative driving applications. In section \RN{3}, we will describe the overall architecture of OpenCDA and reveal the details of each major component. In section \RN{4}, we will showcase a concrete example of how platooning, one of the most important applications in CDA, is implemented under our framework. Afterward, a case study for platooning scenario testings will be introduced. Section \RN{5} will present the experiment results using our default guidance algorithm and compare it with customized algorithms to prove the effectiveness of our framework.

\section{Related Work}

In the past decades, various CDA applications have emerged and imposed significant impacts on ITS. One of the representative applications is the Cooperative Adaptive Cruise Control(CACC), which has been studied extensively, such as \cite{Liu2018, Shladover2015COOPERATIVEAC}. CAVs can form stable strings with short following gaps by utilizing CACC, improving the stability, safety, comfort, and traffic performance in terms of throughput and delay\,\cite{Yi2020}. For freeway traffic, the cooperative merge has been a popular topic as it allows the speed coordination between mainline vehicles and merging vehicles to create qualified gaps for safe merging\, \cite{Ma2020}. Speed harmonization also attracts much attention due to its capability of gradually decreasing upstream traffic speed in a heavily congested area to reduce the stop-and-go traffic and prevent congestion formation\,\cite{ Ghiasi2017SpeedHA, GHIASI2019210, ma2016speed, Ali2013}. Moreover, there are also some CDA applications for intersection control, including vehicular trajectory control\,\cite{Zhou2015}, traffic signal control\,\cite{fENG2015}, and joint control of traffic signal and vehicular trajectories\,\cite{YU201889}.

Although extensive research has been carried out in this field, there are few open-sourced simulation platforms for CDA. Segata \textit{et al.}\,\cite{Segata2014} proposes an extension for Veins\,\cite{5510240} to provide the basic platooning capability.  Recently, SUMO\,\cite{SUMO2018} integrates the Simpla package for basic platoon formations. Wu \textit{et al.} \,\cite{wu2020flow} also present a platform for integrating the traffic simulation and reinforcement learning controllers. However, these platforms only stay at the level of traffic analysis and fail to support full-stack software development and testing for CDA, including perception, planning, decision-making, control, and communication.

The FHWA CARMA Program \,\cite{lochrane2020carma}  has developed software platforms for vehicle and infrastructure and tools for full-scale vehicle software simulation and testing. In collaboration with the CARMA Program, OpenCDA, as an open-source project, makes a unique contribution from the perspective of early-stage development and testing using simulation, enabling users can conveniently conduct both task-specific evaluation (e.g. object detection accuracy) and pipeline-level assessment (e.g. traffic safety) on their customized algorithms.

\section{Overview of OpenCDA}

\begin{figure*}[htbp]
\centering
\includegraphics[width=\textwidth]{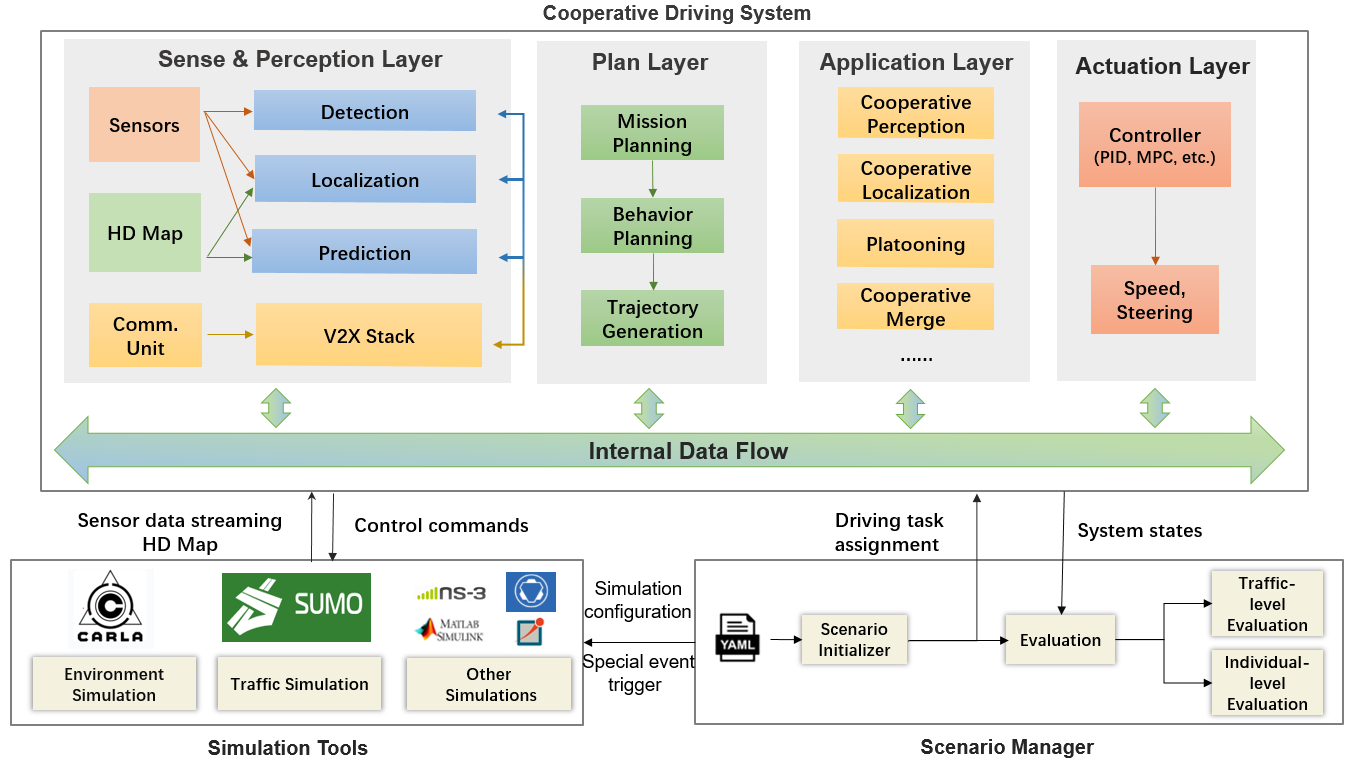}
\caption{The overall architecture design of OpenCDA.\textmd{The full-stack software of the designed cooperative driving system interacts with simulation tools to test the system performance  in provided scenarios}}
\label{fig:overview}
\end{figure*}

OpenCDA is a generalized framework integrated with co-simulation for intelligent and dynamic cooperative driving. It supports various cooperation between automated vehicles and provides benchmarking scenario database and CDA algorithms. As Fig.\,\ref{fig:overview} depicts, OpenCDA is composed of three major components -- the simulation tools, cooperative driving automation system built in Python, and scenario manager. 

\subsection{Simulation Tools}
CARLA\,\cite{Dosovitskiy17} is selected as one of the simulation tools in OpenCDA for automated driving simulation. CARLA is a free, open-source automated driving simulator that aims to accelerate the development of new automated driving technologies. It utilizes Unreal Engine\,\cite{unrealengine} to produce high-quality scene rendering, realistic physics, and basic sensor modeling. The CARLA platform defines a versatile simulation API that users and developers can control over all the elements of the simulation from sensor placement to prototyping and testing the perception, planning, and control algorithms. A key feature of CARLA is its scalable architecture, following a server-multi-client approach to allow for the distribution of computation into multiple nodes. The server will keep updating the physics of the environment, and the client-side will be controlled by users through the CARLA API. Our cooperative driving system is embedded with CARLA API to perform cooperative dynamic driving tasks and evaluate the vehicle performance under the individual autonomy level.

However, CARLA lacks the manageability of large volumes of traffic and fails to represent realistic traffic behavior, thus not ideal for creating a complex traffic environment for CDA testing \,\cite{OlaverriMonreal2018ConnectionOT}. Additionally, CDA's potential in improving overall traffic system performance is also of interest. Therefore, SUMO\,\cite{SUMO2018}, an open-source traffic/driver behavior simulator, is involved in the framework because of its capability of handling large-scale and realistic traffic flows. SUMO has dynamic modeling for each vehicle and allows users to quickly construct customized traffic scenarios through the TraCI (Traffic Control Interface) API. Note that even though CARLA provides a traffic manager module for generating background traffic, they are based on simplistic behavior rules, which cannot represent real driver behavior. 

Further, SUMO can generate traffic using different well-accepted driver models (e.g., Intelligent Driver Model\,\cite{intelligent_drive_model_2010}), and it is more convenient to use SUMO to take in naturalistic trajectory data  (e.g., NGSIM\,\cite{NGGSIM}) and use them directly as the surrounding environment for CDA testing. Since CARLA has developed a co-simulation feature with SUMO, we provide the option for researchers to test their algorithms and protocols solely using CARLA, SUMO,  or employing them together. When the co-simulation is activated,  SUMO will control the traffic and transform the background human-driven vehicles into the CARLA server, and the CAVs controlled by CARLA will react with the traffic to finish their driving tasks. By distributing the tasks to both CARLA and SUMO, the evaluation of designed algorithms or protocols can be processed at both individual level and traffic level.

Note that we do not recommend using the full co-simulation in OpenCDA in all testing tasks. The users need to understand the evaluation needs (e.g., vehicular or traffic behavior) and then select corresponding tools. For example, if only traffic performance is to be understood, there is no point to conduct a fully automated driving simulation and investigate the detailed level of questions such as how sensor outputs and fusion impact traffic performance, because they are not at the same level of analysis. Traffic performance is mostly derived directly from driver/vehicle behavior (as a result of internal mechanisms or algorithms). To this end, in OpenCDA, our benchmark algorithms, to be discussed in the next section, are implemented in both SUMO and CARLA (in a consistent manner, but with some differences due to the fundamental differences between the two simulators in controlling vehicles), so analysis at any level is possible and consistent.

The OpenCDA framework can also be flexibly enhanced with additional tools, such as ns-3 or other customized models for wireless communication. However, we do not consider extremely complex integration of different tools as necessary for the simple reason that no models can fully replicate the real world and models should be built only to meet the testing needs of specific purposes. For example, when evaluating if a certain level of communication packet drops impact traffic stability, using a full ns-3 tool in the simulation loop will not only significantly slow down the simulation but also does not present many benefits as compared to using only simple Monte Carlo simulation\,\cite{hardware_inloop}.

% \begin{figure*}[htbp]
% \centering
%  \includegraphics[width=1.05\linewidth]{images/flow.png} 
% \caption{Logic flow of the cooperative driving system.}
% \label{fig:flo}
% \end{figure*}

\subsection{Cooperative Driving System}
OpenCDA encapsulates the cooperative driving system with CARLA and SUMO via simple API to operate cooperative driving tasks. Sensors mounted on CAVs in CARLA will collect raw sensing information from the simulation environment and proceed to the sensing layer of the system. The received information is then processed by the perception module to perceive the operational environment, utilizing the plan layer to deliver a series of actions, and finally, spawn the control commands through the actuation layer. The actuation will be sent back to CARLA actors to execute the movement at each simulation time step. It is interesting to note that such architecture is also suitable for single-vehicle intelligence development when there is no cooperation needed. This means that the OpenCDA tool can simulate mixed traffic of human-driven, connectivity, and automation.

The cooperation between automated vehicles is activated at the application layer. In this layer, each CAV will exchange status information (e.g., vehicle position, signal phasing, and timing), intent information (e.g., perceived sensing context, planned vehicle trajectory) through the V2X stack, and seek agreement on a plan (e.g., forming a platoon). Based on different agreements, there will be corresponding protocols that potentially modify the default settings of the layers. For instance, when the cooperative perception application is launched, each CAV doesn't solely utilize its own raw sensing information to locate dynamic objects but also retrieves and fuses others' sensing information to achieve multi-modal, cooperative object detection. 

One key feature of the OpenCDA framework is modularity. All layers mentioned above come with default algorithms or protocols, and users can replace the default ones with their customization without influencing others parts by just applying one line of code. We consider this as a desirable feature because researchers can utilize the default modules and algorithms to evaluate the ultimate performance of the entire CDA system and it is also possible for different groups of researchers to compare the algorithms under the same framework. Additionally, the default algorithms in OpenCDA applications, such as cooperative platooning and merge, are also state-of-the-art algorithms that are qualified to serve as benchmark algorithms. Researchers can compare their algorithms with the OpenCDA benchmarks to demonstrate the capability and enhancement of the new algorithms.
 
\subsection{Scenario Manager}
The scenario manager in OpenCDA contains four parts: the scenario configuration file, scenario initializer, special event trigger, and evaluation functions.

A scenario is a description of how the view of the world alters to time. In the context of cooperative driving, it encompasses the information of the static elements of the world (e.g., road topology, surrounding buildings, static objects on the road surface), and dynamic elements such as the traffic flow, traffic signal state, and weather. In OpenCDA, the static elements of a scenario are defined by the default maps in CARLA map library or customized maps built by xdor\,\cite{wiki:xodr} and fbx\,\cite{wiki:fbx} file. The dynamic elements are controlled by a yaml\,\cite{wiki:yaml} file. In the yaml file, users can define the traffic flow for each lane generated by SUMO, including traffic volume and desire speed. If background traffic produced by CARLA is also introduced, then the number and spawn positions of these vehicles and the CARLA traffic manager's settings also need to be recorded. As mentioned before, our framework comes with an existing scenario database that stores predefined scenario testings, but users are welcome to contribute their customized testings to the database.

 After the yaml file is created, a configuration loader will load the file into a Python dictionary. This dictionary will guide the simulation environment construction and determine the major driving tasks for the target CAVs. A driving task composed of the starting locations and destinations of the CAVs, and the intermediate locations to reach.

When the CAVs are executing driving tasks, special events may be triggered. A good example of such events is that a human-driven vehicle in front of a platoon suddenly decelerates or stops. These special events are normally triggered by certain time-step or the positions of CAVs to test the performance of the cooperative driving system in the corner cases. 
% In OpenCDA, the special events need to be defined in an additional OpenSCENARIO format \cite{OpenScenario}.

A driving task is regarded as finished when the CAV arrives at the destination. Then the evaluation is carried out to measure the performance of the whole driving period at an individual level with CARLA and at traffic level with SUMO.

\subsection{Software Class Design and Logic Flow}
To better demonstrate how the interaction of the three major components of OpenCDA are realized, in this section, we will describe the major software class components and the procedure of simulation information transferring between these components by utilizing an example CDA application -- vehicle platooning.

\begin{figure*}[htbp]
\centering
\includegraphics[width=0.8\linewidth]{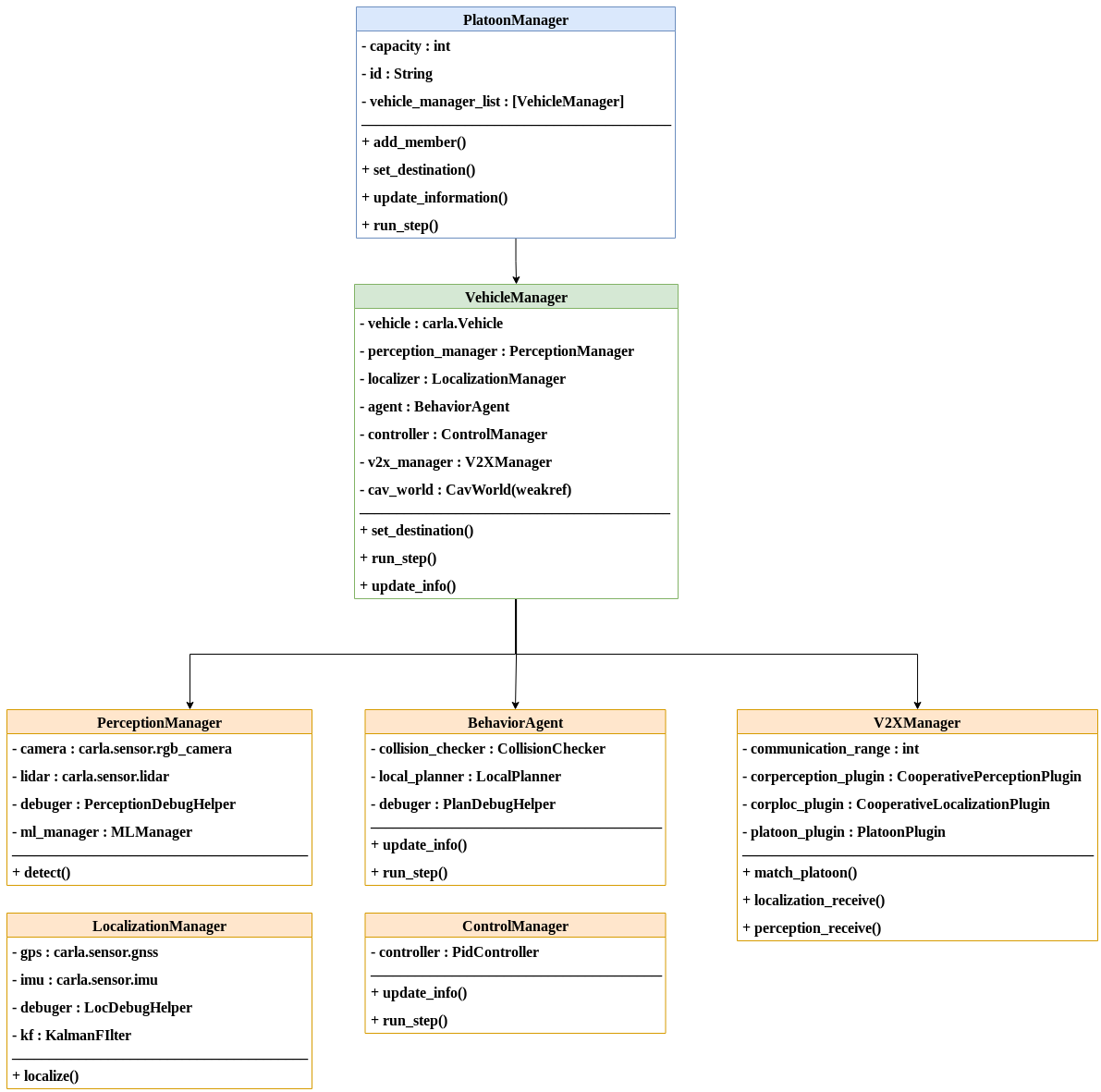}
\caption{Simplified class diagram of platooning. \textmd{Note we only exhibits partial design here.}}
\label{fig:classdiagram}
\end{figure*}

As Fig.\,\ref{fig:classdiagram} depicts, we apply hierarchical class management to control the simulation neatly. The most fundamental class is called \classname{VehicleManager}, which contains the full-stack CDA and Automated Driving System(ADS) benchmark software for a single CAV. The class member \classname{PerceptionManager} and \classname{LocalizationManager} are responsible for perceiving the surrounding environment and localize the ego vehicle. The \classname{BehaviorAgent} plans the driving behavior (e.g. car following, overtaking, lane changing behavior) for the single CAV, and the attribute \classname{LocalPlanner} in \classname{BehaviorAgent} will generate the trajectory using cubic spline interpolation and basic vehicle kinematics:
% \begin{align}
%     y_t= \alpha_0 + \alpha_1x_t + \alpha_2x_t^2 + \alpha_3x_t^3 \\
%     a = 
%     \begin{cases}
%     \min(\frac{v_{target}-v_t}{\Delta{t}}, a^1),& \text{if } v_{target}\geq v_t\\
%         \max(\frac{v_{target}-v_t}{\Delta{t}}, a^2), & \text{otherwise}
% \end{cases} \\
%     x_{t} = v_{t-1}\Delta{t} +\frac{a\Delta{t^2}}{2} \\
%     v_t= v_{t-1} + a\Delta{t}
% \end{align}

\begin{align}
    y_t= \alpha_0 + \alpha_1x_t + \alpha_2x_t^2 + \alpha_3x_t^3 \\
    a_t = 
    \begin{cases}
    \min(\frac{v_{target}-v_t}{\Delta{t}}, a^1),& \text{if } v_{target}\geq v_t\\
        \max(\frac{v_{target}-v_t}{\Delta{t}}, a^2), & \text{otherwise}
\end{cases} \\
    x_{t} = v_{t-1}\Delta{t} +\frac{a_{t-1}\Delta{t^2}}{2} \\
    v_t= v_{t-1} + a_{t-1}\Delta{t}
\end{align}

where $x_t \text{,} ~ y_t$ are the planned x and y coordinates of the vehicle at time step $t$,  $\alpha_0, \alpha_1, \alpha_2$ are the coefficients of cubic polynomial, $a_t$ is the desired acceleration at the time step $t$, $a^1 , a^2$ are the comfort-related acceleration and deceleration, $\Delta{t}$ is the time resolution, i.e., simulation step , $v_{target}, v_t$ are the final target speed and desired speed at time step $t$.  This produced trajectory will be delivered to  \classname{ControlManager}  to generate the throttle, brake, and steering control commands. The \classname{V2XManager} will send and receive the packets (currently regarded as lossless transfer) generated by the components mentioned above to other CAVs for cooperative driving applications.

\begin{figure*}[htbp]
\centering
\includegraphics[width=0.8\linewidth]{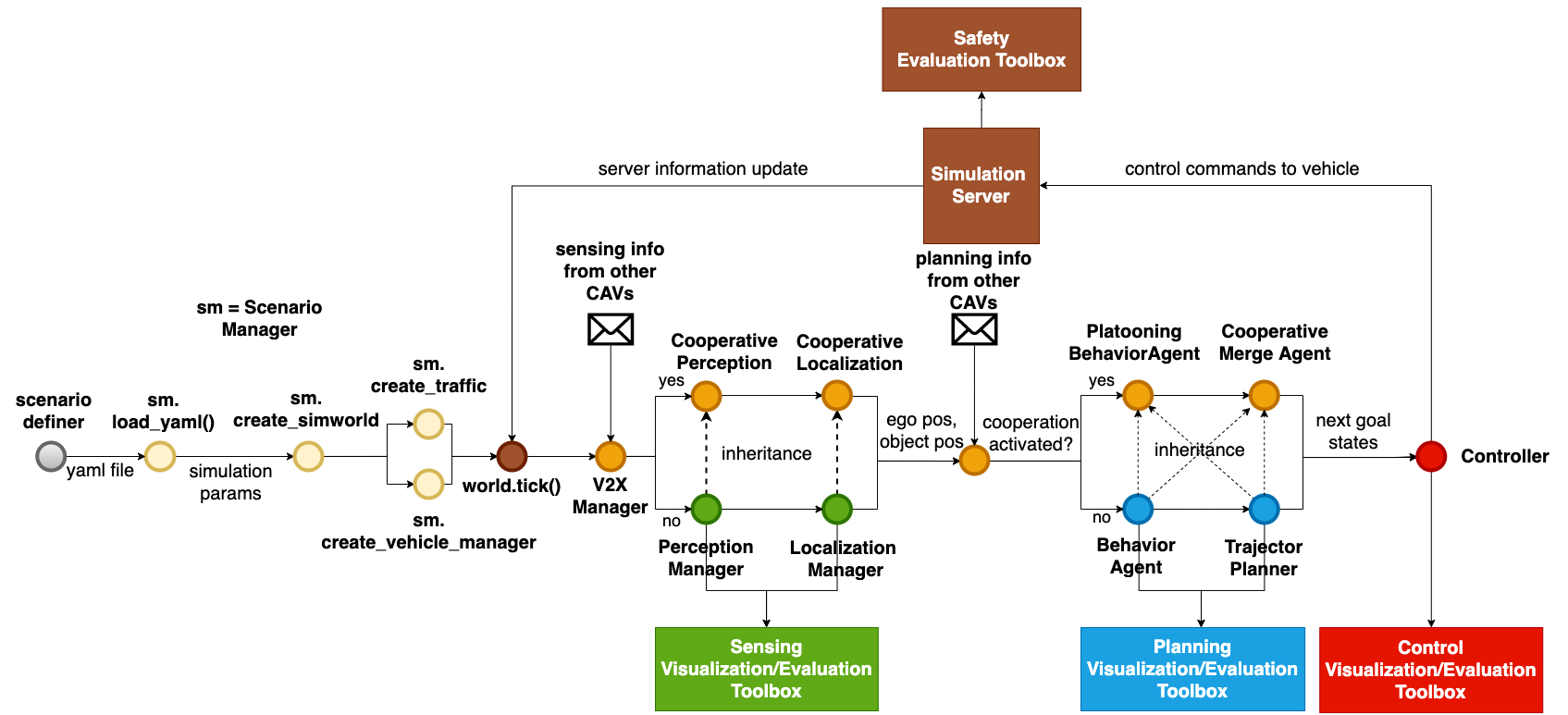}
\caption{Logic flow of the simulation process in OpenCDA.}
\label{fig:flow}
\end{figure*}

Fig.\,\ref{fig:flow} shows the logic flow of the simulation during run time. To run a scenario test, the users are required to create a yaml file based on the template that OpenCDA provides to configure the settings of CARLA server (e.g., synchronous mode versus asynchronous mode), the specifications of the traffic flow (e.g., the number of human drive vehicles, spawn positions), and the parameters of each Connected Automated Vehicle (e.g., sensor parameters, detection model selection, target speed). Subsequently, the \classname{ScenarionManager} will load the configuration file, retrieve the necessary parameters, and deliver them to the CARLA server to settle the simulation environment, generate traffic flow, and create the \classname{VehicleManager} for each CAV. 

After the server updates the information given by \classname{ScenarionManager}, the sensors mounted at each CAV will collect the surrounding environment as well as the ego vehicle information (e.g., 3D LiDAR points, GNSS data) and share those through \classname{V2XManager}. If the upstream cooperative application is activated, \classname{CoopPerceptionManager} and \classname{CoopLocalizationManager} will be utilized to fuse all contexts obtained from other CAVs for object detection and localization. Otherwise, the vehicle will switch to the default \classname{PerceptionManager} and \classname{LocalizationManager}, which do not employ shared data.  The processed sensing information (i.e., object 3D pose, ego position) is delivered to the downstream modules for planning. Similarly, the CAV will select cooperative strategies to make decisions if corresponding applications are activated; otherwise, the origin \classname{BehaviorAgent} and \classname{TrajecotryPlanner} will plan the behavior and generate a smooth trajectory, which is passed to the \classname{ControlManager} to output the final control commands. The CARLA server will apply these commands on the corresponding vehicles, execute a single simulation step, and return the updated information to the \classname{VehicleManager} for the next round of simulation.

It is obvious that the design of the logic flow enhances the flexibility and modularity of OpenCDA as users are capable of choosing the level of cooperation by just modifying the activation indicator. When the simulation is terminated, the embedded evaluation toolboxes will assess the driving performance. We provide default performance measurement for various modules, including perception (e.g., mean average precision of the 3D bounding box detection), localization(e.g. error between estimated and true ego-position), planning (e.g., the smoothness of the planned trajectory), control(e.g, tracking error) and safety (e.g., hazard frequency). If users demand any evaluation measurements that are out of the default scope, they can build customized metrics following the predefined template, which is another key advantage of OpenCDA. 

\section{Experiment Setup and Evaluation Measurement}
\begin{figure*}[htbp]
\centering
\includegraphics[width=\textwidth]{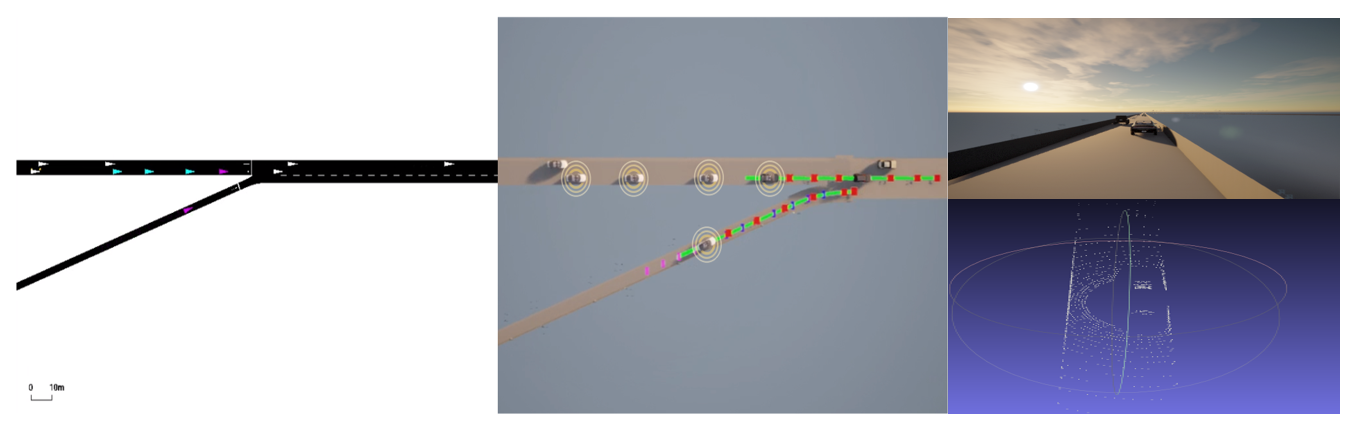}
\caption{A snippet of platooning scenario testing under co-simulation setting. \textmd{From left to right: Sample simulation snippet in SUMO, the corresponding view in CARLA where the green lines and red dots represent planned trajectory path and  points respectively, and the RGB image with 3D lidar points together collected from the sensors mounted at the CAV.}}
\label{fig:platoon_scenrio}
\end{figure*}

To prove the effectiveness of OpenCDA, in this section we continue with an example of vehicle platooning. Our platooning benchmark includes four parts -- the rule-based platooning protocol and algorithms, a customized map with a long freeway basic and merge segment as Fig.\,\ref{fig:platoon_scenrio} presents, several designed testing scenarios, and evaluation metrics. Note that we use such customized map because it leaves enough distance for vehicles to reach high target speed and perform various cooperative maneuvers. For all the experiments conducted, we set the simulation time step as 0.05 second, which means the update frequency of the CARLA server and SUMO is 20Hz.

\subsection{Platooning Protocol Design }
\begin{figure}[htbp]
\centering
\includegraphics[width=\linewidth]{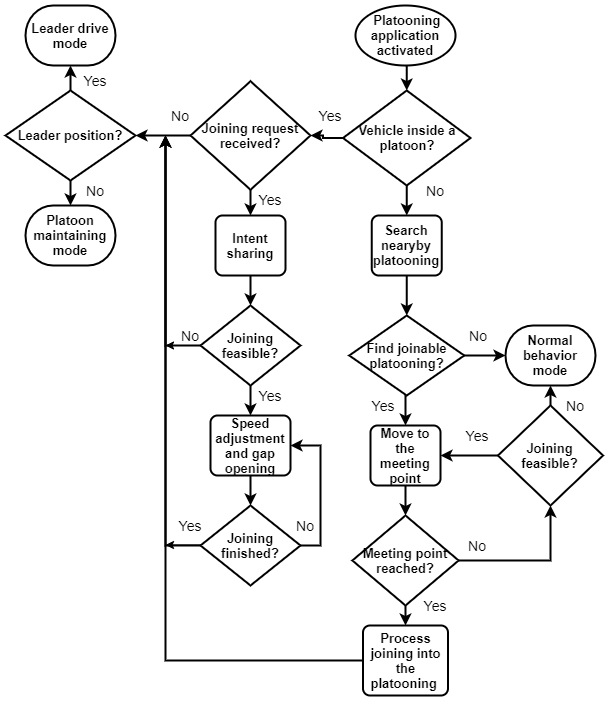}
\caption{Logic Flow of Platooning Protocol.}
\label{fig:platooning}
\end{figure}

As Fig.\,\ref{fig:classdiagram} shows, in the platooning application, all CAVs in the same platoon will be managed by the \classname{PlatoonManager} through a pre-defined protocol. Fig.\,\ref{fig:platooning} displays the default platooning protocol in OpenCDA. Overall, the driving task in a platoon can be divided into different sub-tasks, and platoon members have various driving modes based on the current platooning status.  

When the platooning application is activated, the leading vehicle of the existing platoon will keep listening to the joining requests from CAVs through \classname{V2XManager}. If no such requests are received, the whole platoon will keep moving forward steadily while the leading vehicle will stay in the leader drive mode, in which the vehicle shares a similar behavior pattern with CAVs outside a platoon except overtaking is forbidden. Meanwhile, if the cooperative perception application is also activated, each platoon member will also share its raw sensing information (e.g., camera RGB images, 3d lidar points) and processed sensing information (e.g., detected objects, calibrated vehicle position) retrieved from \classname{PerceptionManager} with the leading vehicle for a better perception.

When there is no joining request, the following vehicles in the platoon will enter the maintaining mode, in which the driving task is defined as adjusting the velocity smoothly to keep a constant inter-vehicular time gap to the preceding members. To accomplish such tasks,  the members need to receive some of the preceding vehicle's trajectory (e.g., leader, immediate predecessor) from \classname{V2XManager} to assist the \classname{LocalPlanner} creating the trajectories, as shown below.

\begin{align}
    pos_j^t = \frac{pos_{j-1}^t - L_{j-1} + pos_{j}^{t-\Delta{t}} \times gap/\Delta{t}}{1 + gap/\Delta{t}} \\
        v_j^t = \frac{||{pos_{j}^t - pos_{j}^{t-\Delta{t}}}||}{\Delta{t}}
\end{align}

where $pos_j^t$, $pos_{j-1}^t$ are the position of vehicle $j$ and its proceeding vehicle $j-1$ at time step $t$, $L_{j-1}$ is the length of vehicle $j-1$, $\Delta{t}$ is the time resolution i.e, simulation step,  $gap$ is the desired inter-vehicular time gap, and $v_j^t$ is the desired speed of the $j$th vehicle at time step $t$. In this example, the platooning algorithm only considers the planned trajectory of the immediate preceding vehicle.
  
If the platoon receives a joining request, the leading vehicle will exchange destination, current position, and planned routes with the requesting CAV to decide whether a feasible joining can be operated. If the request is rejected, the single CAV will keep searching and stay in single-vehicle driver mode. Otherwise, the \classname{PlatoonManager} will choose the best meeting position for the merging vehicle to join depending on the internal and surrounding information, and if needed, certain platoon members will adjust their speed to open a gap for joining. Then the merging vehicle can move to the meeting point and finish the joining maneuver.

\begin{figure}[htbp]
\centering
 \includegraphics[width=1.05\linewidth]{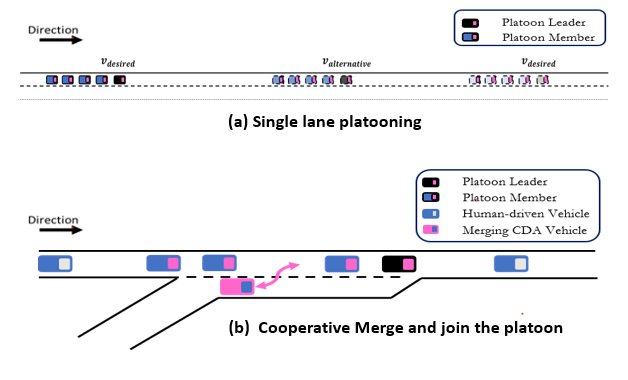} 
\caption{Two different platooning scenario testings.}
\label{fig:4SCENEROS}
\end{figure}

\subsection{ Platooning Scenario Testing Design }
Fig.\,\ref{fig:platoon_scenrio} shows a snippet of the platooning co-simulation testing using the customized benchmark map of a basic freeway merge segment included in OpenCDA. This map is formed by a 2800 meters two-lane freeway for the mainstream traffic and a single-lane on-ramp to allow the merging vehicles to enter the freeway. In this section, we will exhibit two different platooning testing scenarios from our database in this benchmark map. Note that all tests are operated within perception and localization algorithms. We apply yolov5~\cite{glenn_jocher_2020_4154370} for object detection and utilize GNSS/IMU fusion algorithm similar with ~\cite{xia2021autonomous, xiong2020imu} for localization.
 
\subsubsection{Single Lane Platooning}
As Fig.\,\ref{fig:4SCENEROS}(a) describes, there is a five-vehicle platoon that keeps driving in the same lane in this scenario. The objective is to test the platoon's stability, which is indicated by the degree of amplified oscillations when the leading vehicle changes speed dramatically. To meet such a purpose, the platoon leader will follow a given speed profile to accelerate and decelerate frequently to identify whether the following members are capable of maintaining desired inter-vehicular time gap and dampen the speed oscillation. The OpenCDA provides benchmarking testing scenarios of front vehicle trajectories, e.g., two types of cycles of testing with distinct speed profiles. Users can also use their own scenarios for specific purposes by using the example format.

\begin{figure}[htbp]
\centering
\includegraphics[width=0.7\linewidth]{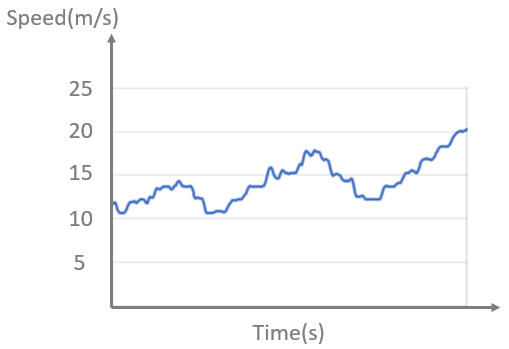}
\caption{Real-world human-driven vehicle speed profile. }
\label{fig:hv}
\end{figure}

 \begin{itemize}
     \item  In the first cycle, the platoon leader will follow a synthetic speed trajectory. It drives at 25 m/s for 20 seconds, then accelerates until reaching the target speed of 30 m/s. The platoon leader will keep this speed for 20 seconds, then decelerate to reach the initial speed of 25 m/s, and keep this speed for 20 seconds. There is no traffic flow generated as we aim to sorely evaluate the platooning protocol in this cycle. Furthermore, the aggressiveness of the acceleration or deceleration and speed maintaining duration can be easily modified to divergent levels, and here we just showcase a single instance.
     
    \item In the second cycle, we placed a human-driven vehicle in front of the platoon. This human-driven vehicle will follow representative speed profiles extracted from NGSIM data, which is collected from real-world experiments as Fig.\,\ref{fig:hv} displays. The leader demands to have a decent car following behavior, and the platoon needs to remain stable while the human-driven vehicle radically adjusts the speed. In such a way, both the car following behaviors of the platoon leader and the platoon followers will be validated. 
 \end{itemize}
 
 \subsubsection{Cooperative Platoon Joining from Other Lanes}
As shown in Fig.\,\ref{fig:4SCENEROS}(b), the mainline has a high-speed traffic flow mixed with human-driven vehicles managed by SUMO and CAVs controlled by CARLA. When the single CAV is near the merging area, it will communicate with the mainline platoon and make a request to join. Once they achieve an agreement, the single CAV has to finish the merge and join the platoon simultaneously before the acceleration lane ends. The leader will decide the best merging position and command certain platoon members to create a gap for the new member. 

To demonstrate OpenCDA's high modularity and extensibility, we further compare two different algorithms of choosing the best merging position. The first approach is heuristic-based. The single CAV will choose the vehicle in the platoon that has the shortest Euclidean distance as the frontal vehicle for merging. The second method is Genetic Fuzzy System \,\cite{Anoop2021}, which utilizes fuzzy logic to decide the best merging position. Different from a heuristic-based method, it also takes platoon members' speed and surrounding human-driven vehicles' information into consideration.

\subsection{Evaluation Measurements}
As adequate performance measurements are essential in the testing, we also provide default evaluation metrics in the scenario benchmark. For platooning application, we assess the performance from \textbf{safety}, \textbf{stability}, and \textbf{efficiency}. 

\subsubsection{Safety}
Safety is always the most critical factor for any automated driving system. In platooning, not only the leading vehicle needs to avoid collisions with surrounding human-driven vehicles, but also the following members are required to keep a safe distance from each other. The safety element can be measured from two perspectives: 
\begin{itemize}
    \item \textbf{Time-to-Collision}:  Time-to-Collision(TTC) refers to the time required for two vehicles to collide if they continue at their present speed and on the same path. Here we can extract the TTC performance series of each vehicle throughout the simulation. It is also possible to estimate the average $TTC$ for each platoon member across all  simulation time steps to represent overall safety by the following  equation:
    \begin{align}
    ATTC = \frac{ \sum_{t=1}^{N}\frac{ x_{i}^t - x_{i-1}^t - l}{ v_{i}^t - v_{i-1}^t} }{N}
    \end{align}
    where $ x_{i}^t $ is the position of vehicle $i$ at time-step $t$, $ x_{i-1}^t $ is the position of the preceding vehicle $i$ at time-step $t$, $l$ is the length of vehicle $i$, $ v_{i}^t,  v_{i-1}^t$ are the speed of vehicle $i$ and $i-1$ at time step $t$ and $N$ is the number of simulation time-steps at which meets the condition $v_{i}^t < v_{i-1}^t$.
    \item \textbf{Hazard frequency}: The number of events that $TTC < TTC_t$, where $TTC_t$ is the warning threshold of Time-to-Collision to  distinguish between safe and unsafe events. In this experiment, we set it as 2.5 second, which is suggested by\,\cite{Bella2011}.

\end{itemize}

\subsubsection{Stability}
The stability of a platoon indicates whether oscillations are amplified from downstream to upstream vehicles \,\cite{Naus2010}. As it is directly correlated with safety and energy consumption, proposing corresponding appropriate evaluation measurements are crucial. In OpenCDA, the following three metrics are used and users can easily define advanced metrics using the data provided by OpenCDA.

\begin{itemize}
    \item \textbf{Inter-vehicular time gap}: The time gap between a platoon member and its proceeding vehicle. The time gap at each simulation step is collected and plotted, and its mean value and standard deviation across the whole episode are calculated as well.
    \item \textbf{Acceleration} The time-series data of acceleration and statistics of the data (e.g., mean, standard deviation) are calculated to reflect the driving smoothness of the platoon members.
\end{itemize}

\begin{table}[]
\begin{subtable}[h]{0.45\textwidth}
\begin{tabular}{l|ll|lll|ll}
\cline{1-6}
\cellcolor[HTML]{C0C0C0} &
  \multicolumn{2}{c|}{\cellcolor[HTML]{C0C0C0}Safety} &
  \multicolumn{3}{l|}{\cellcolor[HTML]{C0C0C0}{\color[HTML]{000000} Stability}} &
  \multicolumn{2}{l}{\cellcolor[HTML]{C0C0C0}Efficiency} \\ \cline{2-8} 
\multirow{-2}{*}{\cellcolor[HTML]{C0C0C0}Vehicle id} &
  attc &
  hf &
  atg &
  tg\_std &
  acc\_std &
  tcm &
  acc\_std \\ \hline
0 & NA    & 0 & NA    & NA    & 0.98 & NA & NA \\ \hline
1 & 30.55 & 0 & 0.603 & 0.007 & 0.73 & NA & NA \\ \hline
2 & 30.50 & 0 & 0.602 & 0.003 & 0.65 & NA & NA \\ \hline
3 & 30.43 & 0 & 0.602 & 0.004 & 0.62 & NA & NA \\ \hline
4 & 30.40 & 0 & 0.602 & 0.005 & 0.60 & NA & NA \\ \hline
\end{tabular}
\caption{Single platooning cycle 1}
\end{subtable}
\label{tab:results_a}

\begin{subtable}[h]{0.45\textwidth}
\begin{tabular}{l|ll|lll|ll}
\cline{1-6}
\cellcolor[HTML]{C0C0C0} &
  \multicolumn{2}{c|}{\cellcolor[HTML]{C0C0C0}Safety} &
  \multicolumn{3}{l|}{\cellcolor[HTML]{C0C0C0}{\color[HTML]{000000} Stability}} &
  \multicolumn{2}{l}{\cellcolor[HTML]{C0C0C0}Efficiency} \\ \cline{2-8} 
\multirow{-2}{*}{\cellcolor[HTML]{C0C0C0}Vehicle id} &
  attc &
  hf &
  atg &
  tg\_std &
  acc\_std &
  tcm &
  acc\_std \\ \hline
0 & 32.68    & 0 & NA    & NA    & 1.42 & NA & NA \\ \hline
1 & 17.1 & 0 & 0.614 & 0.03 & 1.08 & NA & NA \\ \hline
2 & 17.3 & 0 & 0.609 & 0.012 & 0.75 & NA & NA \\ \hline
3 & 18.16 & 0 & 0.608 & 0.007 & 0.55 & NA & NA \\ \hline
4 & 18.92 & 0 & 0.605 & 0.003 & 0.49 & NA & NA \\ \hline
\end{tabular}
\caption{Single platooning cycle 2}
\end{subtable}
\label{tab:results_b}

\begin{subtable}[h]{0.45\textwidth}
\begin{tabular}{l|ll|lll|ll}
\cline{1-6}
\cellcolor[HTML]{C0C0C0} &
  \multicolumn{2}{c|}{\cellcolor[HTML]{C0C0C0}Safety} &
  \multicolumn{3}{l|}{\cellcolor[HTML]{C0C0C0}{\color[HTML]{000000} Stability}} &
  \multicolumn{2}{l}{\cellcolor[HTML]{C0C0C0}Efficiency} \\ \cline{2-8} 
\multirow{-2}{*}{\cellcolor[HTML]{C0C0C0}Vehicle id} &
  attc &
  hf &
  atg &
  tg\_std &
  acc\_std &
  tcm &
  acc\_std \\ \hline
0 & 49.8    & 0 & NA    & NA    & 0.92 & NA & 0.03 \\ \hline
1 & 25.5 & 0 & 0.607 & 0.005 & 0.63 & NA & 0.01 \\ \hline
2 & 31.03 & 0 & 0.607 & 0.003 & 0.56 & NA & 0.01 \\ \hline
3 & 31.53 & 0 & 0.612 & 0.013 & 1.33 & 13.5 & 2.83 \\ \hline
4 & 32.59 & 0 & 0.707 & 0.23 & 0.82 & NA & 1.37 \\ \hline
\end{tabular}
\caption{Cooperative merge and platoon join using heuristic method}
\end{subtable}
\label{tab:results_d}

\begin{subtable}[h]{0.45\textwidth}
\begin{tabular}{l|ll|lll|ll}
\cline{1-6}
\cellcolor[HTML]{C0C0C0} &
  \multicolumn{2}{c|}{\cellcolor[HTML]{C0C0C0}Safety} &
  \multicolumn{3}{l|}{\cellcolor[HTML]{C0C0C0}{\color[HTML]{000000} Stability}} &
  \multicolumn{2}{l}{\cellcolor[HTML]{C0C0C0}Efficiency} \\ \cline{2-8} 
\multirow{-2}{*}{\cellcolor[HTML]{C0C0C0}Vehicle id} &
  attc &
  hf &
  atg &
  tg\_std &
  acc\_std &
  tcm &
  acc\_std \\ \hline
0 & 49.8    & 0 & NA    & NA    & 0.95 & NA & 0.02 \\ \hline
1 & 31.40 & 0 & 0.608 & 0.007 & 1.27 & 9.9 & 2.51 \\ \hline
2 & 31.24 & 0 & 0.674 & 0.16 & 0.79 & NA & 1.18 \\ \hline
3 & 30.14 & 0 & 0.609 & 0.008 & 0.65 & NA & 0.88 \\ \hline
4 & 29.9 & 0 & 0.607 & 0.006 & 0.61 & NA & 0.59 \\ \hline
\end{tabular}
\caption{Cooperative merge and platoon join using GFS}
\end{subtable}
\label{tab:results_e}

\caption{\textbf{Quantitative results of two different scenario tests}. \textmd{The desired platoon time gap is set to 0.6 second. attc:average time-to-collision(second), hf:hazard frequency(number of times), atg:average platoon time gap(second), tg\_std:platoon time gap standard deviation(second), tcm: time to complete maneuver(second)} }
\label{tab:results}
\end{table}

\subsubsection{Efficiency}
The efficiency refers to the time duration required for platoon joining and the smoothness of the joining process. It can be evaluated as follows:
\begin{itemize}
    \item \textbf{Time to complete the maneuver} The time duration starting from the joining request approved to joining concluded.
    \item \textbf{Acceleration} The standard deviation of acceleration during the joining procedure.
    \item \textbf{Traffic delay and throughput} If SUMO are used and traffic performance is of interest, overall delay and throughout of traffic are calculated during the specified simulation period.
\end{itemize}

\begin{figure*}
    \centering
    \begin{subfigure}[b]{\textwidth}
        \centering
        \stackunder[5pt]{\includegraphics[width=\textwidth]{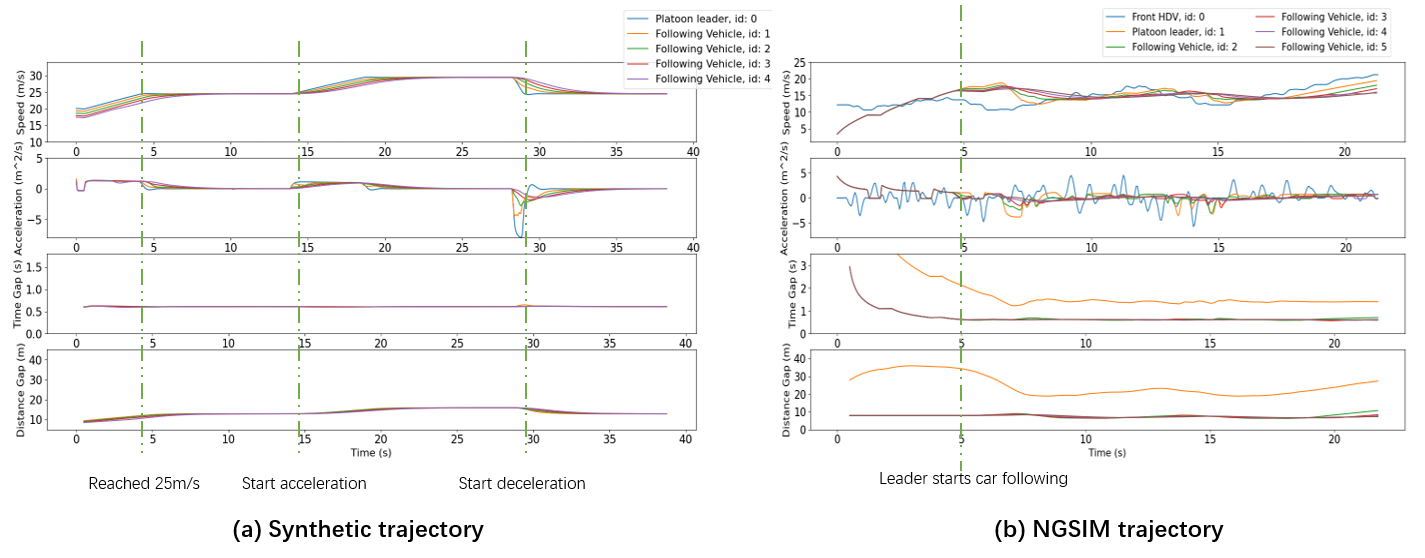}}{}%
        
    \end{subfigure}

    \vskip\baselineskip
    \begin{subfigure}[b]{\textwidth}   
        \centering 
        \includegraphics[width=\textwidth]{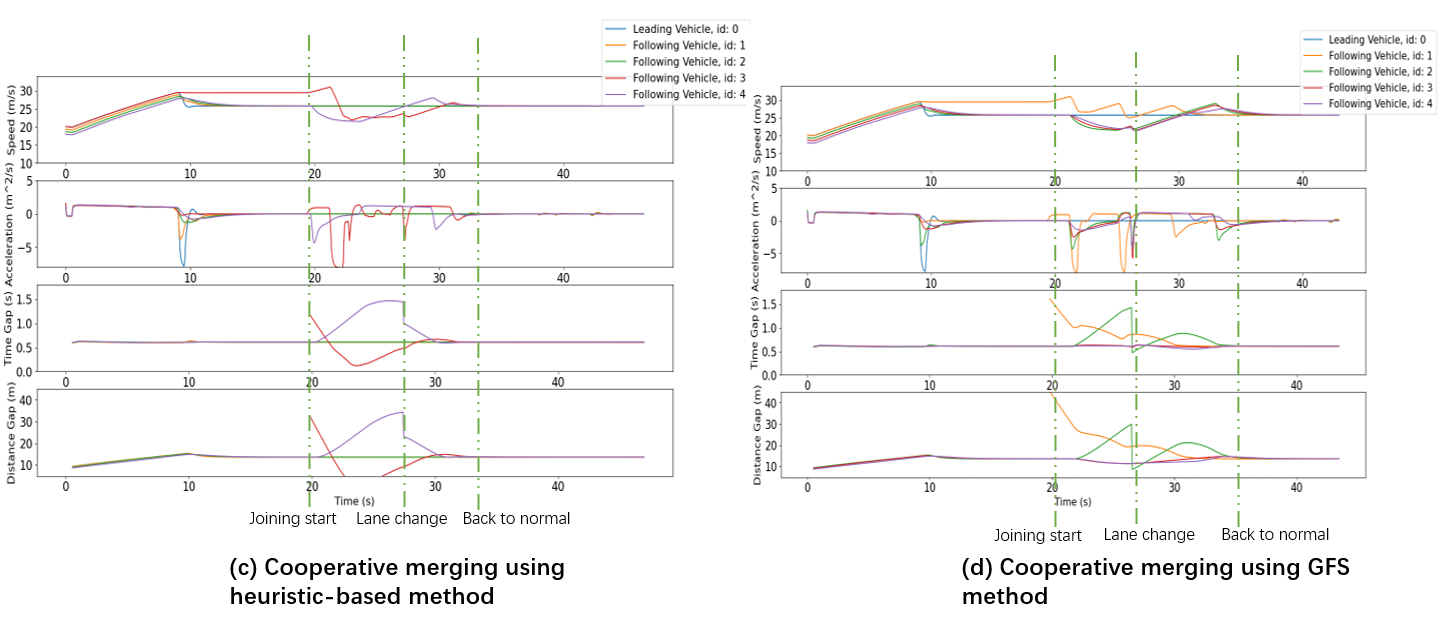}
    \end{subfigure}
    \caption{The speed, acceleration, time gap and distance gap plotting for each CAV in the four testing scenarios}
    \label{fig:plotting}
\end{figure*}

% \begin{figure*}[htbp]
% \centering
% \includegraphics[width=\textwidth]{images/results_two_plot.PNG}
% \caption{The speed, acceleration, time gap and distance gap plotting for each CAV in the second scenario testing.}
% \label{fig:plotting}
% \end{figure*}

\section{Results Analysis}
    In this section, the results for our  benchmark platooning algorithm are presented and discussed.
    
    % Within our pre-installed guidance platooning algorithms, all the tasks assigned in the three scenarios have been accomplished without any collision. 
\subsection{Single Lane Platooning}

Table\,\ref{tab:results}\,(a) presents the average performance of platooning in cycle 1. As we can see, in spite of the dramatic velocity fluctuation, the platoon members can still maintain the desired 0.6 second time gap safely with minor deviations. Similarly, as Table\,\ref{tab:results}\,(b) demonstrates, the leading vehicle is able to follow the human-driven vehicle safely and smoothly while the whole platoon can achieve good safety and stability.  

Fig.\,\ref{fig:plotting} further describes the driving performance at each simulation time step. For the first cycle, as Fig.\,\ref{fig:plotting}(a) demonstrates, the platoon followers are able to keep the designed time gap 0.6s during the whole process, even with the leading vehicle dramatically increasing and decreasing speeds. When the platoon leader starts to accelerate suddenly, the platoon members are able to follow it tightly without any speed-overshooting. When the platoon leader rapidly steps on the brake, the followers can smoothly decelerate at a comfortable rate and stay constant time gaps between each other, which indicates the stability of the platooning. In the real trajectory testing, as Fig.\,\ref{fig:platooning}(b) depicts,  despite the frequent speed changes of the human-driven vehicle, the platoon leader is able to follow it safely, and the time gap between them is around 1.5s. Meanwhile, the platoon member can still keep a constant time gap of 0.6s even when the front human-driven vehicle rapidly accelerates or decelerates. 

These results of the benchmark algorithms illustrate the whole module pipeline of the cooperative driving system in our framework is complete and can work properly for cooperative driving tasks in the simulation environment under various settings. 

\subsection{Cooperative Merge and Join Platoon}
Fig.~\ref{fig:plotting}\,(c) and \ref{fig:plotting}\,(d) display the profiles of velocity, acceleration, inter-vehicular time gap, and distance gap of each of the platoon members during the scenario testing utilizing two different merging position decision algorithms. First, the results of these two algorithms are noticeably distinct. The heuristic-based method chooses the third platoon member as the immediate proceeding vehicle for joining, while the GFS chooses the leading vehicle. Second, when the merging CAV operates the cut-in joining using a heuristic-based method, the time gap between it and the rear member drops under 0.2 seconds, which is potentially dangerous.  In contrast, the GFS allows the merging vehicle to keep the time gap above 0.6 seconds during the whole joining process, which makes the merging process much safer. Last, the GFS is more efficient as Table\ref{tab:results}(e) depicts, it takes 9.9 seconds to end the joining maneuver while the heuristic-based method needs 13.1 seconds. 

In conclusion, the evaluation shows that the GFS is superior to the heuristic-based method. More importantly, our framework allows efficient and straightforward method replacement in as simple as one line of code while maintaining the functionality of the system and the accuracy of other existing modules. This example perfectly proves the effectiveness of OpenCDA in terms of validating any customized CDA algorithms.

\section{Conclusion}
In this article, we introduce OpenCDA, a generalized framework and tool for research and development of Cooperative Driving Automation (CDA). OpenCDA addresses the gap in the community and is one of the first of its kind -- an easy-to-use fast-prototyping tool that has a full-stack CDA software platform that covers perception, communication, planning, and control, to enable researchers to evaluate and compare new CDA algorithms and functions with benchmarks. The six key features of OpenCDA -- \textbf{Connectivity}, \textbf{Integration}, \textbf{Full-stack System}, \textbf{Modularity}, and \textbf{Benchmark} -- have been discussed in detail through the introduction of the OpenCDA architecture, simulation flow, testing scenarios and processes, and software design. By exploiting a practical example of the platooning application, we demonstrate that the modular pipeline in OpenCDA can function properly for CDA applications and the whole framework is flexible enough for any customization. Last, but not least, OpenCDA is an evolving project, and we expect that our team at the UCLA Mobility Lab and interested parties in the community to continuously contribute to the project with additional CDA applications, testing scenarios, enhancements to the existing CDA platform, and integration with other tools for necessary testing purposes.

\bibliographystyle{unsrt}
\bibliography{bio}

\begin{thebibliography}{10}

\bibitem{SAE}
{On-Road Automated Driving (ORAD) committee}.
\newblock Sae j3216 standard: Taxonomy and definitions for terms related to
  cooperative driving automation for on-road motor vehicles.
\newblock In {\em SAE International}, 2020.

\bibitem{6489852}
A.~{Stevens} and J.~{Hopkin}.
\newblock Benefits and deployment opportunities for vehicle/roadside
  cooperative its.
\newblock In {\em IET and ITS Conference on Road Transport Information and
  Control (RTIC 2012)}, pages 1--6, 2012.

\bibitem{lochrane2020carma}
Taylor Lochrane, Laura Dailey, and Corrina Tucker.
\newblock Carma℠: Driving innovation.
\newblock {\em Public Roads}, 83(4), 2020.

\bibitem{hyldmar2019}
Nicholas Hyldmar, Yijun He, and Amanda Prorok.
\newblock A fleet of miniature cars for experiments in cooperative driving.
\newblock In {\em 2019 International Conference on Robotics and Automation
  (ICRA)}, pages 3238--3244, 05 2019.

\bibitem{vanRossum1991InteractivelyTR}
G.~vanRossum and J.~Deboer.
\newblock Interactively testing remote servers using the python programming
  language.
\newblock {\em CWI quarterly}, 4:283--304, 1991.

\bibitem{Dosovitskiy17}
Alexey Dosovitskiy, German Ros, Felipe Codevilla, Antonio Lopez, and Vladlen
  Koltun.
\newblock {CARLA}: {An} open urban driving simulator.
\newblock In {\em Proceedings of the 1st Annual Conference on Robot Learning},
  pages 1--16, 2017.

\bibitem{SUMO2018}
Pablo~Alvarez Lopez, Michael Behrisch, Laura Bieker-Walz, Jakob Erdmann,
  Yun-Pang Fl{\"o}tter{\"o}d, Robert Hilbrich, Leonhard L{\"u}cken, Johannes
  Rummel, Peter Wagner, and Evamarie Wie{\ss}ner.
\newblock Microscopic traffic simulation using sumo.
\newblock In {\em The 21st IEEE International Conference on Intelligent
  Transportation Systems}. IEEE, 2018.

\bibitem{ns32017}
George~F. Riley and Thomas~R. Henderson.
\newblock The ns-3 network simulator.
\newblock In Klaus Wehrle, Mesut Günes, and James Gross, editors, {\em
  Modeling and Tools for Network Simulation}, pages 15--34. Springer, 2010.

\bibitem{carsim}
{Mechanical Simulation }.
\newblock Carsim, 2021.

\bibitem{Liu2018}
Liu Hao, Xingan David~Kan Lin~Xiao, Xiao-Yun~Lu Steven E.~Shladover,
  Wouter~Schakel Meng~Wang, and Bart van Arem.
\newblock Using cooperative adaptive cruise control (cacc) to form
  high-performance vehicle streams.final report.
\newblock 2018.

\bibitem{Shladover2015COOPERATIVEAC}
S.~Shladover, C.~Nowakowski, Xiaoyun Lu, and Robert~E. Ferlis.
\newblock Cooperative adaptive cruise control (cacc) definitions and operating
  concepts.
\newblock 2015.

\bibitem{Yi2020}
Yi~Guo and Jiaqi Ma.
\newblock Leveraging existing high-occupancy vehicle lanes for mixed-autonomy
  traffic management with emerging connected automated vehicle applications.
\newblock {\em Transportmetrica A: Transport Science}, 16(3):1375--1399, 2020.

\bibitem{Ma2020}
Jiaqi Ma, Edward Leslie, Amir Ghiasi, and Yi~Guo.
\newblock Empirical analysis of a freeway bundled connected-and-automated
  vehicle application using experimental data.
\newblock {\em Journal of Transportation Engineering}, 146, 04 2020.

\bibitem{Ghiasi2017SpeedHA}
A.~Ghiasi, Jiaqi Ma, Fang Zhou, and Xiaopeng Li.
\newblock Speed harmonization algorithm using connected autonomous vehicles.
\newblock 2017.

\bibitem{GHIASI2019210}
Amir Ghiasi, Xiaopeng Li, and Jiaqi Ma.
\newblock A mixed traffic speed harmonization model with connected autonomous
  vehicles.
\newblock {\em Transportation Research Part C: Emerging Technologies},
  104:210--233, 2019.

\bibitem{ma2016speed}
Jiaqi Ma, Xiaopeng Li, Steven Shladover, Hesham~A. Rakha, Xiao-Yun Lu,
  Ramanujan Jagannathan, and Daniel~J. Dailey.
\newblock Freeway speed harmonization.
\newblock {\em IEEE Transactions on Intelligent Vehicles}, 1(1):78--89, 2016.

\bibitem{Ali2013}
Alireza Talebpour, Hani~S. Mahmassani, and Samer~H. Hamdar.
\newblock Speed harmonization: Evaluation of effectiveness under congested
  conditions.
\newblock {\em Transportation Research Record}, 2391(1):69--79, 2013.

\bibitem{Zhou2015}
Fang Zhou, Xiaopeng Li, and Jiaqi Ma.
\newblock Parsimonious shooting heuristic for trajectory design of connected
  automated traffic part i: Theoretical analysis with generalized time
  geography.
\newblock {\em Transportation Research Part B Methodological}, 95, 11 2015.

\bibitem{fENG2015}
Yiheng Feng, Larry Head, Shayan Khoshmagham, and Mehdi Zamanipour.
\newblock A real-time adaptive signal control in a connected vehicle
  environment.
\newblock {\em Transportation Research Part C: Emerging Technologies}, 55, 01
  2015.

\bibitem{YU201889}
Chunhui Yu, Yiheng Feng, Henry~X. Liu, Wanjing Ma, and Xiaoguang Yang.
\newblock Integrated optimization of traffic signals and vehicle trajectories
  at isolated urban intersections.
\newblock {\em Transportation Research Part B: Methodological}, 112:89--112,
  2018.

\bibitem{Segata2014}
Michele Segata, Stefan Joerer, Bastian Bloessl, Christoph Sommer, Falko
  Dressler, and Renato Lo~Cigno.
\newblock Plexe: A platooning extension for veins.
\newblock volume 2015, 12 2014.

\bibitem{5510240}
C.~{Sommer}, R.~{German}, and F.~{Dressler}.
\newblock Bidirectionally coupled network and road traffic simulation for
  improved ivc analysis.
\newblock {\em IEEE Transactions on Mobile Computing}, 10(1):3--15, 2011.

\bibitem{wu2020flow}
Cathy Wu, Aboudy Kreidieh, Kanaad Parvate, Eugene Vinitsky, and Alexandre~M
  Bayen.
\newblock Flow: A modular learning framework for autonomy in traffic, 2020.

\bibitem{unrealengine}
{Epic Games}.
\newblock Unreal engine.

\bibitem{OlaverriMonreal2018ConnectionOT}
C.~Olaverri-Monreal, Javier Errea-Moreno, Alberto D{\'i}az-{\'A}lvarez, Carlos
  Biurrun-Quel, Luis Serrano-Arriezu, and Markus Kuba.
\newblock Connection of the sumo microscopic traffic simulator and the unity 3d
  game engine to evaluate v2x communication-based systems.
\newblock {\em Sensors (Basel, Switzerland)}, 18, 2018.

\bibitem{intelligent_drive_model_2010}
Arne Kesting, Martin Treiber, and Dirk Helbing.
\newblock Enhanced intelligent driver model to access the impact of driving
  strategies on traffic capacity.
\newblock {\em Philosophical Transactions of the Royal Society A: Mathematical,
  Physical and Engineering Sciences}, 368(1928):4585--4605, 2010.

\bibitem{NGGSIM}
U.S.~Department of~Transportation Federal Highway~Administration.
\newblock Next generation simulation (ngsim) vehicle trajectories and
  supporting data, 2016.

\bibitem{hardware_inloop}
Jiaqi Ma, Fang Zhou, Zhitong Huang, Christopher Melson, Rachel James, and
  Xiaoxiao Zhang.
\newblock Hardware-in-the-loop testing of connected and automated vehicle
  applications: A use case for queue-aware signalized intersection approach and
  departure.
\newblock {\em Transportation Research Record Journal of the Transportation
  Research Board}, 2672, 01 2018.

\bibitem{wiki:xodr}
{OpenDRIVE}.
\newblock Opendrive --- {W}ikipedia{,} the free encyclopedia, 2005.

\bibitem{wiki:fbx}
{FBX}.
\newblock Fbx file format --- {W}ikipedia{,} the free encyclopedia, 2006.

\bibitem{wiki:yaml}
{YAML}.
\newblock Yaml --- {W}ikipedia{,} the free encyclopedia, 2001.

\bibitem{glenn_jocher_2020_4154370}
Glenn Jocher, Alex Stoken, Jirka Borovec, NanoCode012, ChristopherSTAN, Liu
  Changyu, Laughing, tkianai, Adam Hogan, lorenzomammana, yxNONG, AlexWang1900,
  Laurentiu Diaconu, Marc, wanghaoyang0106, ml5ah, Doug, Francisco Ingham,
  Frederik, Guilhen, Hatovix, Jake Poznanski, Jiacong Fang, Lijun Yu,
  changyu98, Mingyu Wang, Naman Gupta, Osama Akhtar, PetrDvoracek, and Prashant
  Rai.
\newblock {ultralytics/yolov5: v3.1 - Bug Fixes and Performance Improvements},
  October 2020.

\bibitem{xia2021autonomous}
Xin Xia, Ehsan Hashemi, Lu~Xiong, Amir Khajepour, and Nan Xu.
\newblock Autonomous vehicles sideslip angle estimation: Single antenna
  gnss/imu fusion with observability analysis.
\newblock {\em IEEE Internet of Things Journal}, 2021.

\bibitem{xiong2020imu}
Lu~Xiong, Xin Xia, Yishi Lu, Wei Liu, Letian Gao, Shunhui Song, and Zhuoping
  Yu.
\newblock Imu-based automated vehicle body sideslip angle and attitude
  estimation aided by gnss using parallel adaptive kalman filters.
\newblock {\em IEEE Transactions on Vehicular Technology}, 69(10):10668--10680,
  2020.

\bibitem{Anoop2021}
Anoop Sathyan, Jiaqi Ma, and Kelly Cohen.
\newblock Decentralized cooperative driving automation: A reinforcement
  learning framework using genetic fuzzy systems.
\newblock {\em Transportmetrica B}, 07 2021.

\bibitem{Bella2011}
Francesco Bella and Roberta Russo.
\newblock A collision warning system for rear-end collision: a driving
  simulator study.
\newblock {\em Procedia - Social and Behavioral Sciences}, 20:676--686, 12
  2011.

\bibitem{Naus2010}
Gerrit Naus, Rene Vugts, Jeroen Ploeg, M.J.G. Molengraft, and M.~Steinbuch.
\newblock String-stable cacc design and experimental validation: A
  frequency-domain approach.
\newblock {\em Vehicular Technology, IEEE Transactions on}, 59:4268 -- 4279, 12
  2010.

\end{thebibliography}

\end{document}